\pdfoutput=1

\documentclass[11pt]{article}

\usepackage{acl}
\usepackage{soul}
\usepackage{times}
\usepackage{latexsym}

\usepackage[T1]{fontenc}

\usepackage[utf8]{inputenc}

\usepackage{microtype}

\usepackage{inconsolata}
\usepackage{graphicx}
\usepackage{hyperref}
\usepackage{booktabs}
\usepackage{xcolor}
\usepackage{colortbl}
\usepackage{tabularx}
\usepackage{multirow}
\usepackage{tikz}
\usepackage{pgfplots}
\pgfplotsset{compat=1.18} 
\definecolor{lightgray}{gray}{0.95}
\definecolor{lightgray2}{gray}{0.9}
\definecolor{grass}{RGB}{0, 148, 50}

\newcommand{\declarelogo}[0]{\includegraphics[height=.02\textwidth]{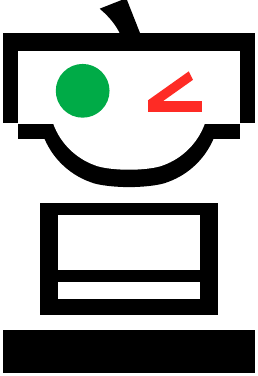}}

\title{Contrastive Chain-of-Thought Prompting}

\author{
\textbf{
Yew Ken Chia\thanks{~~Equal contribution. Yew Ken and Guizhen are students under the Joint PhD Program between Alibaba and their corresponding university. 
}
~\textsuperscript{\rm 1,~${\declarelogo}$}
\quad
Guizhen Chen\footnotemark[1]\textsuperscript{\rm ~~~1,~2}
} \\
\textbf{
Luu Anh Tuan\textsuperscript{\rm 2} 
\quad
Soujanya Poria\textsuperscript{\rm ${\declarelogo}$}
\quad 
Lidong Bing\thanks{~~Corresponding author.}\textsuperscript{\rm ~~~1}
} \\
\textsuperscript{\rm 1}DAMO Academy, Alibaba Group, Singapore~~
\textsuperscript{\rm ${\declarelogo}$} Singapore University of Technology and Design ~~\\
\textsuperscript{\rm 2}Nanyang Technological University, Singapore \\
{\tt\{yewken\_chia, sporia\}@sutd.edu.sg} \\
{\tt\{guizhen001, anhtuan.luu\}@ntu.edu.sg} \\
{\tt\{yewken.chia, guizhen.chen, l.bing\}@alibaba-inc.com}
\\ 
}

\begin{document}
\maketitle
\begin{abstract}
Despite the success of chain of thought in enhancing language model reasoning, the underlying process remains less well understood.
Although logically sound reasoning appears inherently crucial for chain of thought, prior studies surprisingly reveal minimal impact when using invalid demonstrations instead.
Furthermore, the conventional chain of thought does not inform language models on what mistakes to avoid, which potentially leads to more errors.
Hence, inspired by how humans can learn from both positive and negative examples, we propose contrastive chain of thought to enhance language model reasoning.
Compared to the conventional chain of thought, our approach provides both valid and invalid reasoning demonstrations, to guide the model to reason step-by-step while reducing reasoning mistakes.
To improve generalization, we introduce an automatic method to construct contrastive demonstrations.
Our experiments on 
reasoning benchmarks demonstrate that contrastive chain of thought can serve as a general enhancement of chain-of-thought prompting.\footnote{Our code implementation will be released at \href{https://github.com/DAMO-NLP-SG/contrastive-cot}{
https://github.com/DAMO-NLP-SG/contrastive-cot}}

\end{abstract}

\begin{figure}[!t]
\centering
\includegraphics[width=1.0\columnwidth]{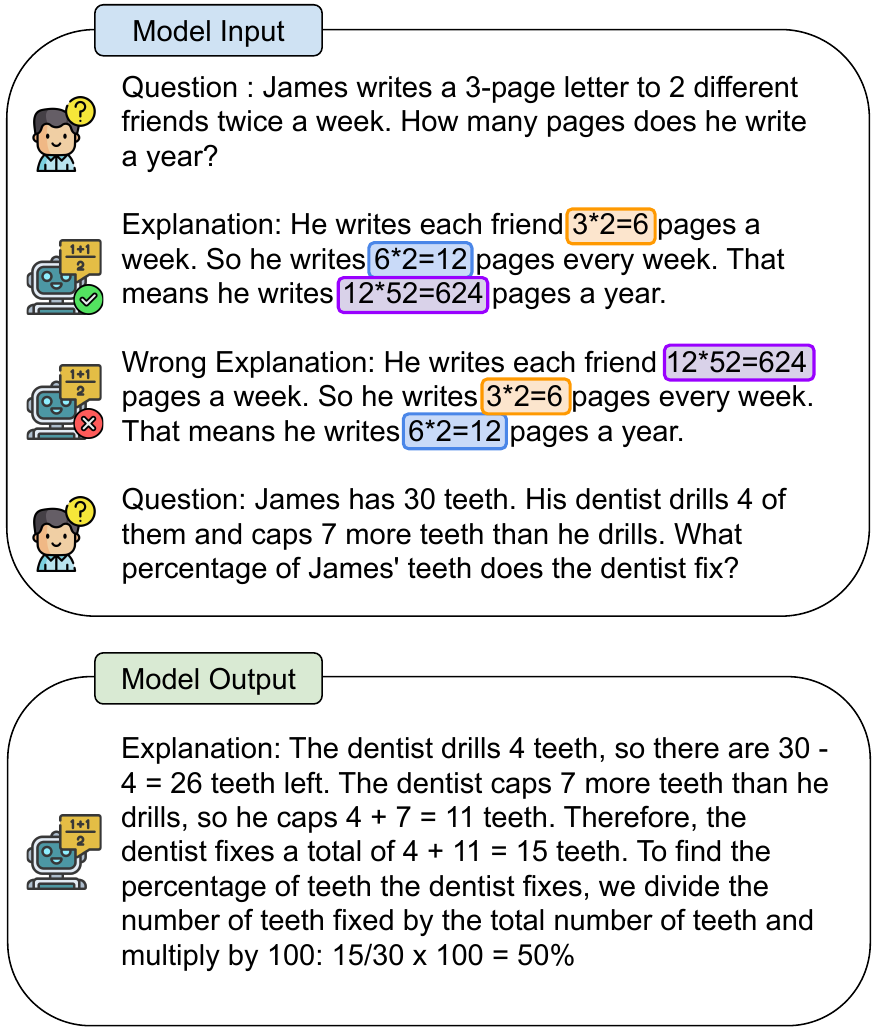}
\caption{Example of contrastive chain-of-thought which leverages both positive and negative demonstrations to enhance language model reasoning.}
\label{fig:cover}
\end{figure}

\section{Introduction}


With the trend of large language models (LLMs), massively scaling the model size has enabled greater generalization \cite{NEURIPS2020_1457c0d6} and the emergent ability to perform new tasks when given suitable prompts \cite{Wei2022EmergentAO}.
However, solely increasing the model size cannot solve complex reasoning tasks \cite{gopher}.
To this end, chain-of-thought prompting was proposed to unlock the reasoning ability of LLMs by generating intermediate reasoning steps \cite{wei2022chain}. 
In practice, most methods based on chain of thought leverage in-context learning \cite{NEURIPS2020_1457c0d6}by prompting the model with demonstrations of the input, chain-of-thought, and output \cite{cot_survey}.


However, despite its success, we lack a thorough understanding of the chain of thought \cite{DBLP:journals/corr/abs-2101-09194}.
For example, it was shown that even demonstrations with invalid reasoning can lead to similar performance compared to valid demonstrations \cite{wang-etal-2023-towards}\footnote{Note that while chain-of-thought can be performed in a zero-shot fashion with prompts, we focus on the few-shot setting, as it was originally proposed in \citet{wei2022chain}.}. 
Hence, it is not clear how language models learn to reason effectively based on the chain-of-thought demonstrations. 
On the other hand, 
mistakes in the intermediate steps can compound and derail the reasoning process \cite{ling2023deductive}.
Any potential error in the reasoning process not only affects the accuracy of the final result but also undermines the trustworthiness of the language model \cite{turpin2023language}.
Thus, it is also important to reduce mistakes in intermediate reasoning steps.

\begin{figure*}[!t]
\centering
\includegraphics[width=1.0\linewidth]{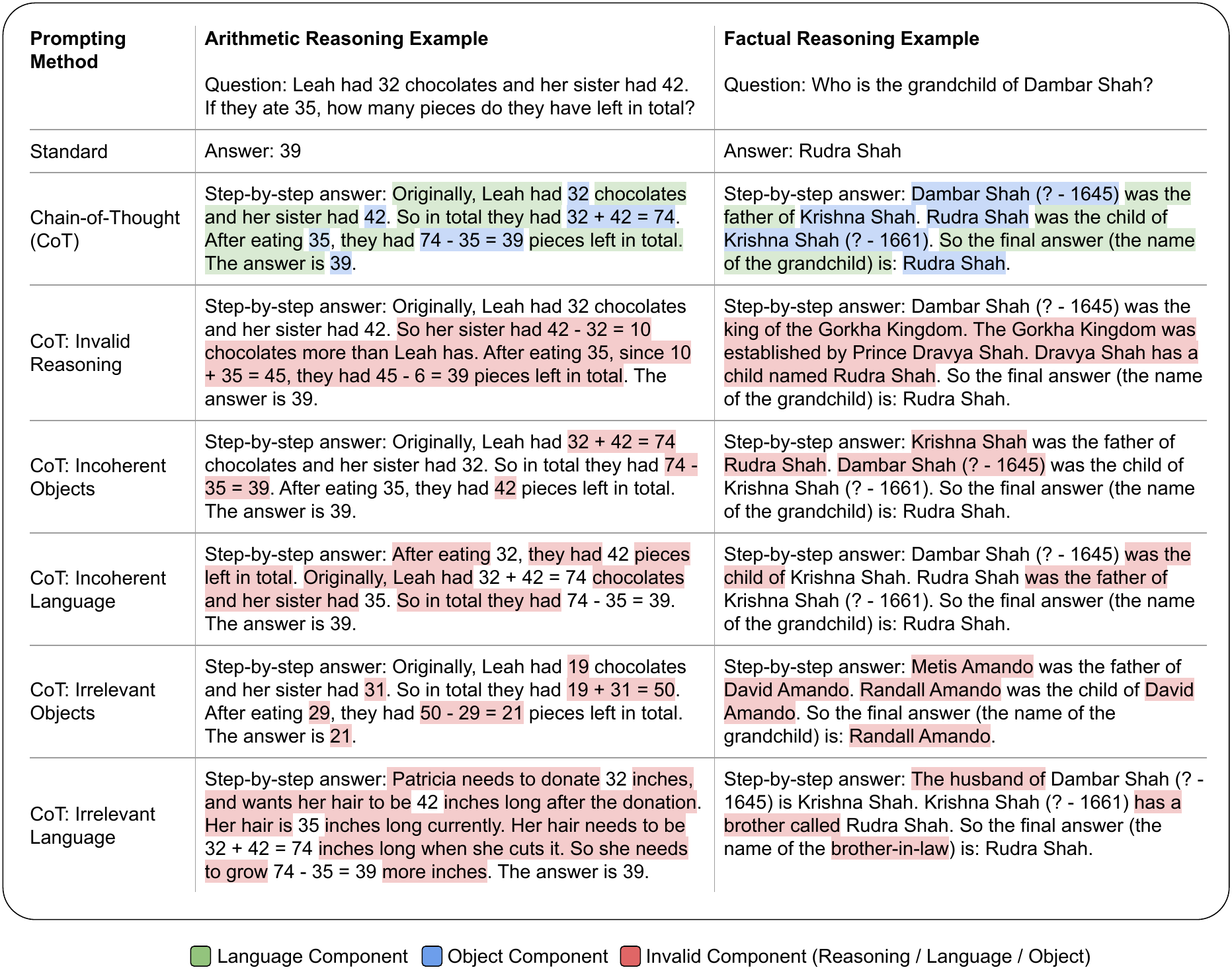}
\caption{Categorization of invalid chain-of-thought examples, following \citet{wang-etal-2023-towards}.}
\label{fig:intro}
\end{figure*}

To address the challenges of chain of thought, we are inspired by how humans can learn from positive as well as negative examples.
For instance, when solving a complex task where the intermediate steps are not well-defined, it is useful to learn the correct steps from positive demonstrations, as well as avoiding faults in negative demonstrations.
Hence, we propose contrastive chain of thought, which provides both positive and negative demonstrations to enhance the reasoning of language models.
Naturally, this raises the question of how to design effective negative demonstrations, as well as whether they can be generalized to diverse tasks.
Through our analysis of multiple invalid reasoning types, we design a simple and effective method that can automatically generate contrastive demonstrations from existing valid reasoning chains.
Furthermore, as contrastive chain-of-thought is task-agnostic and compatible with methods such as self-consistency \cite{Wang2022SelfConsistencyIC}, we believe that it can serve as a general enhancement of chain of thought.


To measure the effectiveness of contrastive chain of thought, we present evaluations on a wide range of reasoning benchmarks, and find significant benefits.
Notably, compared to conventional chain of thought, we observe improvements of 9.8 and 16.0 points for GSM-8K \cite{gsm8k} and Bamboogle \cite{bamboogle} respectively when using GPT-3.5-Turbo\footnote{\href{https://platform.openai.com/docs/models}{https://platform.openai.com/docs/models}}, a widely used LLM.
Further analysis of the reasoning chains generated from our method also shows significant reduction in errors.

In summary, our main contributions include: 
(1) We analyse various invalid reasoning types and find that combining positive and negative demonstrations generally boost the effectiveness of chain-of-thought.
(2) Based on the analysis above, we propose contrastive chain of thought to enhance language model reasoning. To improve generalization, we also propose an automatic method to construct contrastive demonstrations.
(3) Evaluations on multiple reasoning benchmarks demonstrate significant improvements compared to conventional chain of thought.


\section{Preliminary Study: Effect of Different Types of Contrastive Demonstrations}
\label{sec:pre_study}
While chain of thought (CoT) prompting has enhanced the reasoning of large language models, it remains less well understood.
For instance, while sound reasoning seems intuitively important to effective chain of thought, previous work has shown that there is little effect when using invalid demonstrations.
On the other hand, previous works in contrastive learning \cite{NEURIPS2020_d89a66c7} and alignment \cite{ouyang2022training} have demonstrated how language models can learn more effectively from both valid and invalid examples.
Hence, we conduct a preliminary study with the following research question: \textbf{Can invalid reasoning demonstrations be instead used to enhance chain of thought?}
Specifically, we aim to study the effect of providing chain-of-thought demonstrations in a ``contrastive'' manner, i.e., demonstrations containing both valid and invalid rationales.

\subsection{Components of Chain of Thought}
Compared to standard prompting with in-context demonstrations \cite{NEURIPS2020_1457c0d6}, chain-of-thought (CoT) prompting \cite{wei2022chain} includes a rationale for each demonstration example.
Each rationale consists of a series of intermediate reasoning steps, guiding the language model to solve tasks in a step-by-step manner.
Following the formulation of \cite{wang-etal-2023-towards}, we identify two distinct components of each CoT rationale:
\begin{itemize}
    \item \textcolor{cyan}{Bridging objects} are the symbolic items that the model traverses in order to reach the final solution. 
    For example, the objects could be numbers and equations in arithmetic tasks, or the names of entities in factual tasks.
    \item \textcolor{grass}{Language templates} are the textual hints that guide the language model to derive and contextualize the correct bridging objects during the reasoning process.
\end{itemize}

\subsection{What is Invalid Chain of Thought?}
Given the distinct components of chain of thought, we are now able to systematically identify the aspects which lead to invalid rationales.
Concretely there are two main aspects which are applicable to both the language and object components:

\begin{itemize}
    \item \textbf{Coherence} refers to the correct ordering of steps in a rationale, and is necessary for successful chain of thought.
    Specifically, as chain of thought is a sequential reasoning process, it is not possible for later steps to be pre-conditions of earlier steps.
    \item \textbf{Relevance} refers to whether the rationale contains corresponding information from the question. 
    For instance, if the question mentions a person named Leah eating chocolates, it would be irrelevant to discuss a different person cutting their hair.
\end{itemize}

In addition, following \citet{wang-etal-2023-towards}, we include invalid reasoning as a category of invalid chain of thought, which is neither incoherent nor irrelevant, but contains logical mistakes.
Hence, we aim to study the five main categories of invalid chain-of-thought, as shown in Figure \ref{fig:intro}.

\subsection{Experimental Setup}
\label{sec:setup}
To conduct the experiments for the preliminary study, we leverage the GSM8K \cite{gsm8k} and Bamboogle \cite{bamboogle} datasets for arithmetic and factual reasoning respectively.
We use the OpenAI Chat Completions API\footnote{\href{https://platform.openai.com/docs/api-reference}{https://platform.openai.com/docs/api-reference}} which is one of the most popular and well-performing language models with reasonable cost.
Specifically, we use the GPT-3.5-Turbo (0301) version.
To study the effect of contrastive demonstrations under various settings, we evaluate the five main invalid categories as shown in Figure \ref{fig:intro}.
Note that we use 4-shot prompting for each dataset, and the chain-of-thought demonstrations are manually constructed by previous works \cite{wei2022chain, wang-etal-2023-towards}.
To standardize the prompting process, we use a simplified chain-of-thought prompt format, as shown in Figure \ref{fig:cover}.

\subsection{Preliminary Results}



\begin{figure*}[!t]
\centering
\includegraphics[width=1.0\linewidth]{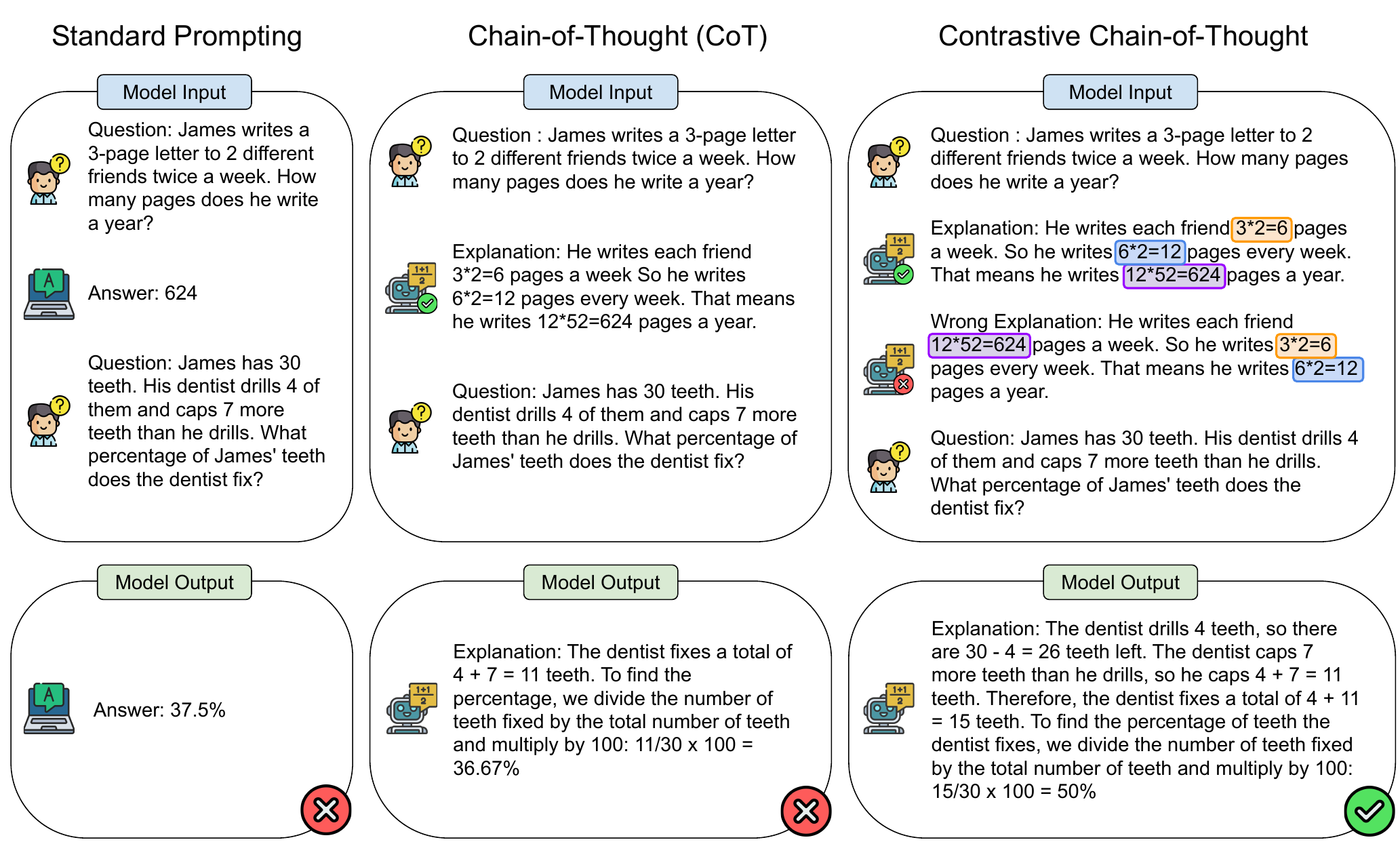}
\caption{Overview of contrastive chain-of-thought (right), with comparison to common prompting methods.}
\label{fig:compare}
\end{figure*}

\begin{table}
    \centering
    \resizebox{1\columnwidth}{!}{
    \begin{tabular}{lccc}
    \toprule
    \textbf{Prompting Method} &\textbf{GSM8K} &\textbf{Bamboogle} & \textbf{Avg.}\\
    \midrule
    Standard & 27.4 & 11.2 & 19.3 \\
    Chain-of-Thought &69.2 &40.8 &55.0 \\
    \quad w/ Invalid Reasoning &76.0 &45.6 &60.8\\
    \quad w/ Incoherent Objects &79.6 &\textbf{53.6} &\textbf{66.6}\\
    \quad w/ Incoherent Language &78.8 &52.8 &65.8\\
    \quad w/ Irrelevant Objects &79.8 &48.8 &64.3\\
    \quad w/ Irrelevant Language &\textbf{80.2} &49.6 &64.9 \\
    \bottomrule
    \end{tabular}
}
    \caption{Preliminary results on the effect of contrastive demonstrations for chain of thought.}
    \label{tab:preliminary}
\end{table}

\begin{table*}[!t]
    \centering
    \resizebox{1.0\linewidth}{!}{
    \begin{tabular}{llllllll}
    \toprule
    \multirow{2}{*}{\textbf{Prompting Method}} &\multicolumn{5}{c}{\textit{Arithmetic Reasoning}} 
    &\multicolumn{2}{c}{\textit{Factual QA}}\\
    \cmidrule(lr){2-6}
    \cmidrule(lr){7-8}
    &\textbf{GSM8K} 
    &\textbf{AQuA} 
    &\textbf{GSM-Hard} 
    &\textbf{SVAMP} 
    &\textbf{ASDIV} 
    &\textbf{Bamboogle} 
    &\textbf{StrategyQA} \\
    \midrule
    Standard & 27.4 & 29.5 & 11.2 & 69.3 & 75.8 & 12.0 & 59.4\\
    \rowcolor{lightgray} CoT &69.2 &53.5 &33.8 &67.2 &70.8 &40.8 &55.8 \\
    \rowcolor{lightgray2} Contrastive CoT &79.0 \textcolor{grass}{(+9.8)} 
    &57.5 \textcolor{grass}{(+3.9)} 
    &44.2 \textcolor{grass}{(+10.4)} 
    &81.6 \textcolor{grass}{(+14.4)}
    &84.4 \textcolor{grass}{(+13.6)}
    &56.8 \textcolor{grass}{(+16.0)}
    &66.2 \textcolor{grass}{(+10.4)}\\
    \midrule
    Standard-SC & 28.0& 29.9 & 11.0 & 69.0 & 76.0 & 11.2 & 59.6\\
    \rowcolor{lightgray} CoT-SC &71.0 &55.9 &34.0 &71.6 &74.0 &40.8 &57.0 \\
    \rowcolor{lightgray2} Contrastive CoT-SC &86.2 \textcolor{grass}{(+15.2)}
    &71.7 \textcolor{grass}{(+15.7)}
    &50.0 \textcolor{grass}{(+16.0)}
    &85.2 \textcolor{grass}{(+13.6)}
    &89.6 \textcolor{grass}{(+15.6)}
    &58.4 \textcolor{grass}{(+17.6)}
    &69.6 \textcolor{grass}{(+12.6)}\\
    \bottomrule
    \end{tabular}
    }
    \caption{Main evaluation results for contrastive chain-of-thought on several reasoning tasks.}
    \label{tab:main}
\end{table*}

\begin{table}[t]
    \centering
    \resizebox{1\columnwidth}{!}{
    \begin{tabular}{lccc}
    \toprule
    \textbf{Dataset} & \textbf{Type} & \textbf{$|$Train$|$} & \textbf{$|$Test$|$}\\
    \midrule
    GSM8K & Arithmetic Reasoning & 4 & 500 \\
    AQuA & Arithmetic Reasoning & 4 & 254 \\
    GSM-Hard & Arithmetic Reasoning & 4 & 500 \\
    SVAMP & Arithmetic Reasoning & 4 & 500 \\
    ASDIV & Arithmetic Reasoning & 4 & 500 \\
    Bamboogle & Factual QA & 4 & 125 \\
    StrategyQA & Factual QA & 4 & 500 \\
    \bottomrule
    \end{tabular}
}
    \caption{Details of datasets used.}
    \label{tab:data}
\end{table}

Based on the preliminary results in Table \ref{tab:preliminary}, we observe significant gains across all invalid rationale categories compared to conventional chain-of-thought.
Notably, leveraging chain of thought with contrastive demonstrations containing incoherent objects yields
the highest average performance on GSM8K and Bamboogle.
This suggests that language models are better able to learning step-by-step reasoning when provided with both valid and invalid rationales.
Hence, we believe that contrastive demonstrations have the potential to greatly enhance language model reasoning ability.

\section{Contrastive Chain of Thought}
Chain-of-thought (CoT) prompting, as evidenced by prior research, has indeed elevated the reasoning capabilities of large language models \cite{wei2022chain}. 
However, a comprehensive understanding of this phenomenon is still lacking.
Although logically sound reasoning appears to be inherently crucial for chain of thought, prior studies surprisingly reveal minimal impact when employing invalid demonstrations.
To this end, based on our preliminary study in Section \ref{sec:pre_study}, we found that providing both valid and invalid reasoning demonstrations in a ``contrastive'' manner greatly improves reasoning performance.
However, this approach may not generalize well to new tasks, as it requires manual construction of the invalid rationales.

Thus, we propose a general prompting method known as contrastive chain of thought, which includes automatic construction of contrastive demonstrations.
Figure \ref{fig:compare} presents an overview of our approach.
Specifically, the language model is provided with the question, ground truth answer explanation and incorrect answer explanation.
Compared to standard prompting, our method enables models to perform more complex reasoning by decomposing problems into intermediate steps. Compared to conventional chain-of-thought prompting, our method contrasts the valid and invalid answer explanations, guiding the model to generate more accurate reasoning chains.

Concretely, given a small set of $n$ in-context demonstration examples $D = \{E_1, \ldots, E_{|n|}\}$, and a query $Q$, the goal of the model is to generate a suitable answer $A$.
For standard prompting, the demonstration examples consist of just the question and answer, i.e., $E_j = (Q_j, A_j)$.
On the other hand, chain-of-thought is a more advanced prompting method that guides the model with intermediate reasoning steps $T$. As shown in the figure above, the reasoning steps $T$ typically consist of multiple sentences where each sentence describes one reasoning step. Hence, chain-of-thought prompting examples consist of the question, reasoning steps, and final answer, i.e., $E_j = (Q_j, T_j, A_j)$. 
However, the model does not know what faults to avoid in conventional chain-of-thought, which could lead to increased mistakes and error propagation.
Hence, our contrastive chain of thought method provides both the correct and incorrect reasoning steps in the demonstration examples, i.e., $E_j = (Q_j, T_{j,+}, A_{j,+}, T_{j,-}, A_{j,-})$. 

To obtain the correct reasoning steps $T_{+}$ for the demonstration examples, we use the annotated examples from the previous chain-of-thought works. 
For the incorrect reasoning steps $T_-$, we automatically construct it from the correct reasoning steps $T_{+}$, based on the "Incoherent Objects" category in Section \ref{sec:pre_study}.
Concretely, we use an existing entity recognition model\footnote{\url{https://spacy.io/models/en\#en_core_web_trf}} to extract the object spans such as numbers, equations, or persons from a given chain-of-thought rationale.
Consequently, we randomly shuffle the position of the objects within the rationale, thus constructing a rationale with incoherent bridging objects.
Note that when testing with a new question, only the question and demonstration examples are provided to the model, and the model must generate its own reasoning steps before producing the final answer.




\section{Experiments}
\subsection{Experimental Setup}
We focus our study on two main types of reasoning tasks: arithmetic reasoning and factual question answering (QA). 
For arithmetic reasoning, we conduct experiments on a range of datasets including GSM8K \cite{gsm8k}, AQuA \cite{aqua}, GSM-Hard \cite{pmlr-v202-gao23f}, SVAMP \cite{svamp}, and ASDIV \cite{asdiv}. 
For factual QA, we include two datasets: Bamboogle \cite{bamboogle} and StrategyQA \cite{strategyqa}. 
To maintain a reasonable computing budget, we limit each dataset to a maximum of 500 test samples through random sampling.
For datasets that contain less than 500 test samples, we instead use all available test samples.
The datasets' details are included in Table \ref{tab:data}.
Regarding model and prompting details, we use the same experimental setup as for our preliminary study in Section \ref{sec:pre_study}.



\subsection{Main Results}

To assess the effectiveness of our method, we evaluate on several reasoning tasks and report the main results in Table \ref{tab:main}. 
Our main findings are as follows:

\paragraph{Contrastive CoT demonstrates consistent improvements over conventional CoT.}
Contrastive CoT consistently outperforms conventional CoT across the datasets in both arithmetic and factual reasoning categories.
Notably, we observe substantial gains of more than 10 points on GSM-Hard, SVAMP, ASDIV, Bamboogle and StrategyQA.
Thus, the consistent and significant performance improvements demonstrate the general effectiveness of our proposed method.
As contrastive chain of thought can be automatically constructed from existing rationales, the annotation cost is the same as conventional chain of thought.
Hence, it can be viewed as a general enhancement of chain of thought.

\paragraph{Contrastive CoT is more effective when applied with self-consistency.}
As self-consistency \cite{Wang2022SelfConsistencyIC} is a popular decoding strategy to boost the chain-of-thought performance of large language models, we are interested to see if contrastive chain of thought can benefit similarly from self-consistency. 
In general, we observe that self-consistency further enhances the performance of contrastive CoT. This enhancement is particularly evident in the case of the AQuA dataset. While contrastive CoT alone results in a modest performance improvement of 4.0\%, applying self-consistency amplifies this gain significantly, achieving an additional improvement of 14.2\%.

\section{Related Work}

\paragraph{Large Language Models}
Recent developments in large language models have shown that massively scaling the size and training data of models can greatly improve generalization \cite{DBLP:journals/corr/abs-2001-08361}.
Notably, large language models have been shown to generalize to new tasks when given suitable prompts and demonstrations \cite{NEURIPS2020_1457c0d6}.
This has brought about a new paradigm of leveraging language models for tasks without the need for additional training \cite{10.1145/3560815}.
However, simply scaling language models has not been sufficient to attain good performance on challenging tasks such as arithmetic reasoning and factual question answering \cite{wei2022chain}.
Hence, in this work, we focus on enhancing the reasoning ability of large language models through prompts.

\paragraph{Chain of Thought}
Chain-of-thought prompting was introduced by \citet{wei2022chain} to enhance language model reasoning by generating intermediate steps.
Notably, this has inspired numerous works that build upon this direction of step-by-step reasoning.
For instance, automatic chain-of-thought \cite{zhang2023automatic} was proposed to address the challenges in manually annotating chain-of-thought demonstrations.
On the other hand, it was shown that specific prompts such as ``Let's think step-by-step'' can enable language models to perform chain-of-thought in a zero-shot manner, without any demonstrations \cite{kojima2022large}.
In addition, challenging problems can be decomposed into multiple sub-problems \cite{zhou2023leasttomost}, or even into code programs that can be automatically executed \cite{pmlr-v202-gao23f}.
Despite the progress in chain-of-thought on multiple fronts, we still lack a rigorous understanding of the underlying mechanism \cite{turpin2023language, feng2023towards}.
In this work, inspired by the findings of previous works regarding invalid demonstrations, we propose contrastive chain-of-thought to enhance language model reasoning.
As contrastive chain-of-thought leverages both valid and invalid reasoning demonstrations, we believe this may encourage other researchers to fundamentally rethink the chain-of-thought process.

\paragraph{Learning from Negative Examples}
While chain-of-thought prompting typically involves only valid demonstrations, it is not clear whether invalid demonstrations can also benefit the reasoning process \cite{wang-etal-2023-towards}.
On the other hand, learning from negative or invalid samples is not new.
For instance, contrastive learning is a well-established deep learning approach that encourages models to distinguish between ``positive'' and ``negative'' samples, thus learning better representations \cite{NEURIPS2020_d89a66c7}.
Similarly, reinforcement learning from human feedback (RLHF) trains a reward model based on positive and negative samples of human preference data \cite{ouyang2022training, NIPS2017_d5e2c0ad}.
Hence, inspired by the previous approaches, we propose contrastive chain-of-thought, a general enhancement of chain-of-thought prompting, by enabling models to learn from both valid and invalid reasoning demonstrations.

\section{Conclusions}
In this work, we have explored the effect of leveraging invalid reasoning demonstrations for enhancing chain of thought.
Through our preliminary study on different invalid chain-of-thought categories, we found that providing both valid and invalid demonstrations in a contrastive manner greatly improves reasoning ability in language models.
To overcome the challenge of manually annotating invalid rationales, we propose contrastive chain of thought, a general prompting method which can automatically construct contrastive demonstrations from existing rationales.
Through experiments on several reasoning tasks, we find contrastive chain of thought to be a general enhancement of chain-of-thought prompting.
Further investigation into alternative forms of chain-of-thought prompting will hopefully inspire future advancements in language-based reasoning.


\clearpage
\newpage

\bibliography{custom}




\end{document}